\documentclass[sigconf]{acmart}
\renewcommand\footnotetextcopyrightpermission[1]{} 

\AtBeginDocument{%
  \providecommand\BibTeX{{%
    \normalfont B\kern-0.5em{\scshape i\kern-0.25em b}\kern-0.8em\TeX}}}

\usepackage{subcaption}
\usepackage{array}
\usepackage{booktabs}
\usepackage{multirow}
\usepackage{colortbl}
\usepackage{spreadtab}
\usepackage{longtable}
\usepackage{pdflscape}
\usepackage{float}

\title{Learning to Perceive in Deep Model-Free Reinforcement Learning}

\author{Gonçalo Querido}
\affiliation{
  \institution{Instituto Superior Técnico}
  \city{Lisbon}
  \country{Portugal}}
\email{goncalo.querido@tecnico.ulisboa.pt}

\author{Alberto Sardinha}
\affiliation{
  \institution{INESC-ID \& Instituto Superior Técnico \& PUC-Rio}
  \city{Lisbon}
  \country{Portugal}}
\email{jose.alberto.sardinha@tecnico.ulisboa.pt}

\author{Francisco S. Melo}
\affiliation{
  \institution{INESC-ID \& Instituto Superior Técnico}
  \city{Lisbon}
  \country{Portugal}}
\email{fmelo@inesc-id.pt}

\begin{abstract}
    This work proposes a novel model-free Reinforcement Learning (RL) agent that is able to learn how to complete an unknown task having access to only a part of the input observation. We take inspiration from the concepts of visual attention and active perception that are characteristic of humans and tried to apply them to our agent, creating a hard attention mechanism. In this mechanism, the model decides first which region of the input image it should look at, and only after that it has access to the pixels of that region. Current RL agents do not follow this principle and we have not seen these mechanisms applied to the same purpose as this work. In our architecture, we adapt an existing model called {\em recurrent attention model} (RAM) and combine it with the {\em proximal policy optimization} (PPO) algorithm. We investigate whether a model with these characteristics is capable of achieving similar performance to state-of-the-art model-free RL agents that access the full input observation. This analysis is made in two Atari games, Pong and SpaceInvaders, which have a discrete action space, and in CarRacing, which has a continuous action space. Besides assessing its performance, we also analyze the movement of the attention of our model and compare it with what would be an example of the human behavior. Even with such visual limitation, we show that our model matches the performance of PPO+LSTM in two of the three games tested.

\end{abstract}


\newcommand{\BibTeX}{\rm B\kern-.05em{\sc i\kern-.025em b}\kern-.08em\TeX}

\begin{document}

\pagestyle{plain} 
\maketitle

\section{Introduction} \label{sec:intro}
    In our everyday lives, even though we are constantly being flooded with visual stimuli, we do not give the same importance to everything in our field of view. Instead, we focus on small regions that attract us the most. In those moments, we take advantage of a cognitive process called \textit{visual attention} \cite{connor2004visual}. Unconsciously, we interpret those regions and extract meaning from them using another mental process named \textit{perception} \cite{schacter2016psychology}. The combination of both these processes allows us to solve complex tasks because, from all the visual information we receive, we filter the most important to perform our activities and not pay attention to irrelevant elements in our surroundings.


Current Reinforcement Learning (RL) models, even though they achieve excellent performance in a broad range of tasks, do not follow this behavior typical of humans. For example, when learning to play a video game, RL algorithms typically process the whole input image, giving the same importance to every region of the input game frame. Such design results in models that rely on large convolutional neural networks (CNNs) that process a large number of pixels, making the model take too long to train, requiring high computational power, and potentially limiting their applicability \cite{katharopoulos2019processing}. To keep the training time reasonable, images are often preprocessed to reduce the size of the input, losing some of its details. Using these low-resolution images can hamper the models from completely understanding what is present in their input, which lowers their performance \cite{thambawita2021impact}.

To overcome these limitations, in this paper we contribute the first RL architecture that implements an attention mechanism similar to the one humans have. Applying such a mechanism allows the model to only process the pixels it perceives as the most useful, which makes it much more computationally efficient and able to use the original images without resizing them.

Attention mechanisms such as the one described above recently started appearing in the literature, but none of them were applied to the same purpose as this work. The closest to our work is, perhaps, the model proposed by Mnih et al., called {\em recurrent attention model} (RAM) \cite{mnih2014recurrent}, which implements the same attention mechanism we use in our work in the context of image classification. The authors introduce the concept of \textit{glimpse}, a retina-like representation of a portion of an image centered around a location $l$. The region of the image around $l$ has high resolution; regions further away from $l$ have increasingly lower resolution. Such  representation is crucial to the performance of the agent and is what makes the complexity of the model not dependent on the size of the input images. Instead, it depends on the size of the glimpses. 
The RAM paper uses four networks: one responsible for extracting the glimpses from an image and learning features from them; a Long Short Term Memory (LSTM) network that builds an internal representation of the environment combining the information from all the previous glimpses; one network responsible for the classification of the image received, and another that chooses the coordinates of the next glimpse.

There is also a number of other models that apply different types of attention mechanisms. For example, the work of Tang et al. \cite{tang2020neuroevolution} is an example of an RL agent that is capable of achieving top performance in games such as CarRacing \cite{carRacing2016} and DoomTakeCover \cite{doomTakeCover2016}. Their agent starts by dividing the input video frame into patches, assigning a level of importance to each one of them. Then, it selects the $K$ most important patches and extracts features from them. This information is later used by a controller which selects the actions to take. Although this model only has around 3600 parameters, since it is trained using a computational intensive evolutionary strategy called Covariance-Matrix Adaptation Evolution Strategy (CMA-ES), it takes a long time to train.

In this paper, we address the following research questions:
\begin{enumerate}
    \item Is it possible to attain state-of-the-art performance in complex control tasks with limited (but active) perception?
    \item Is there any similarity between the attention movements of the model and human behavior?   
\end{enumerate}
To address these questions, the paper proposes a novel architecture that combines a glimpse-based attention mechanism with a model-free reinforcement learning algorithm (PPO). Our results show our model can match the performance of PPO+LSTM in two of the three games tested while processing a significantly smaller number of pixels from the input images. Our model has fewer training parameters than its vanilla counterparts and is, therefore, more efficient than existing models that apply attention mechanisms. 

In the remainder of this paper, we go over the modifications we have made to the RAM \cite{mnih2014recurrent} architecture, showing how our model compares against multiple versions of the PPO algorithm when playing video games such as Pong, SpaceInvaders, and CarRacing. In the end, we analyze the movement of the attention of our model and compare it with what would be an example of human behavior.

\section{Background}
    This section introduces the core concepts and notation used in the remainder of the document.

\subsection{Active Perception and Attention}

The concept of {\em active perception} was defined by Bajcsy as the intelligent acquisition of information about an environment in order to understand it better \cite{bajcsy1988active}. In comparison to a passive perception agent that statically senses its environment and takes actions accordingly, an active perception agent can improve its performance by actively moving its sensors to better reason about its environment \cite{lee2021active}. 

In artificial intelligence, active perception models can be implemented using deep learning methods such as CNNs \cite{lee2021active}. Since the selection of what an agent should sense can take inspiration from the human visual attention process, an attention mechanism can be used to replicate such behavior. In machine learning, \textit{attention} is a technique that consists in choosing which parts of the input are the most important, resulting in more computational power being allocated to them. In the literature \cite{ghaffarian2021effect,wang2016survey,yang2020overview}, we can find two categories of attention mechanisms: 

\begin{itemize}
    \item \textbf{Soft attention}: is a mechanism that splits the input into multiple parts, assigning a weight to each one of them. The weights represent the importance associated with each portion of the image, and since this model is differentiable, they can be updated using backpropagation.
    \item \textbf{Hard attention}: is a mechanism where the model does not go through some parts of its input because the neural network itself stochastically decides which are the parts it should pay attention to. However, this mechanism is not differentiable but can be trained using RL.
\end{itemize}

The model from Tang et al. \cite{tang2020neuroevolution} used a soft attention mechanism, while in our work, we use a hard attention mechanism. 

\subsection{Reinforcement Learning} \label{sec:back_rl}

In RL, an agent interacts with its environment and learns, by trial and error, an action-selection rule (a {\em policy}) that maximizes the agent's reward over time. 



In our problem, the agent does not have full observability of the environment, so the best mathematical framework to formally represent this problem is a Partially Observable Markov Decision Process (POMDP). A POMDP is defined as a tuple $(\mathcal{S}, \mathcal{A}, \mathcal{Z}, \mathcal{P}, \mathcal{O}, r)$, where $\mathcal{S}$ is a discrete set of states, $\mathcal{A}$ is a discrete set of actions, $\mathcal{Z}$ is a discrete set of observations, $\mathcal{P}(s' \, | \, s, a)$ represents the transition probability of going from state $s$ to $s'$ performing action $a$, $\mathcal{O}(z \, | \, s', a)$ represent the probability of receiving the observation $z$, while being in state $s'$ after taking action $a$, and $r(s, a)$ is the reward function $r : \mathcal{S} \times \mathcal{A} \rightarrow \mathbb{R}$.

In our work, we decided to propose a model-free RL architecture instead of a model-based. We did not choose the latter because we decided to understand first if an RL model was capable of having good performance with such visual restriction while having a simpler, model-free approach. Building a complete model of the environment while just seeing a region of it, is another challenge that we leave for future work.  
Since we made that decision, our agent is not able to know either the transition probabilities $\mathcal{P}$ or the reward function $r$ of the environment. Therefore, it has to learn, explicitly by trial and error, the optimal policy $\pi: \mathcal{H} \rightarrow \Delta(\mathcal{A})$, which is a mapping from the history of past observations to a distribution over actions. One way of learning that policy is using an actor-critic policy-gradient method. This algorithm learns a parameterized policy (\textit{an actor}) and estimates a value function (\textit{a critic}). For our problem, we need two actors: one to learn the action policy $\pi_{\theta}(a \, | \, z)$ (Actor), and the other to learn the policy $\pi_{\mu}(l \, | \, z)$ that chooses the coordinates of the image to look at (Locator).

The specific actor-critic method used in this work was the PPO algorithm, presented by Schulman et al. \cite{schulman2017proximal}. PPO is simpler than other policy gradient methods, well-studied, and was already tried alongside the RAM architecture \cite{zuur2019deep}. For these reasons, we decided to have it as the base of our model.

In PPO, we alternate between interacting with the environment to get data and updating the policy using stochastic gradient ascent. In every policy update, this policy gradient method guarantees that the difference between the new policy and the old policy is small, which prevents the algorithm from having a high variance during training. Having the probability ratio

\begin{equation}
    r_{t}(\theta) = \frac{\pi_{\theta}(a_{t} \, | \, s_{t})}{\pi_{\theta_{\mathrm{old}}}(a_{t} \, | \, s_{t})}
\end{equation}

and $\hat{A}_{t}$, an estimator of the advantage function at timestep $t$, PPO maximizes the following surrogate objective:

\begin{equation}
    \label{eqn:ppo_clip}
    L^{CLIP}(\theta) = \hat{\mathbb{E}}_{t} \Big[ \min(r_{t}(\theta)\hat{A}_{t}, \mathrm{clip}(r_{t}(\theta), 1 - \epsilon, 1 + \epsilon)\hat{A}_{t}) \Big]
\end{equation}

where $\epsilon$ is a hyperparameter. The clipping prevents large policy updates and penalizes the probability ratio $r_{t}(\theta)$  when it tries to move far away from 1.

\section{Glimpse-Based Actor-Critic (GBAC)}
    In this section, we introduce a novel model called Glimpse-Based Actor-Critic (GBAC) that combines a hard attention mechanism with a model-free RL algorithm.  When compared to other RL models, GBAC processes much fewer pixels, and its training parameters do not depend on the size of the input, which makes it more efficient. 

In our problem, the game environment $E$ gives, at each timestep, a frame $s_{t}$ of the game. Since our model cannot have access to all the information of $s_{t}$, it has to select only a portion of it to be its observation $z_{t}$. That observation has the coordinates $l_{t-1} = (x_{t-1}, y_{t-1})$ that were chosen by the model in the previous timestep. During its interaction with $E$, our agent has to learn the best policy $\pi_{\theta}(a_{t} \, | \, z_{t})$ that selects the action $a_{t}$ to be performed in the game. While doing this, the model also has to understand which regions of $s_{t}$ have the most valuable information and learn a policy $\pi_{\mu}(l_{t} \, | \, z_{t})$ to choose the set of coordinates $(x_{t}, y_{t})$ to be taken in the next timestep.

As we have seen in Section \ref{sec:back_rl}, this problem can be seen as an instance of a POMDP. In our case, the observations the agent takes are represented by the glimpses, and the history of past interactions with the environment is stored in the hidden state of two LSTMs. When combining the memory, the current glimpse, and the previously chosen location, the LSTMs have all the information needed to learn the policies that select the actions and the locations to take next.

\subsection{Architecture}

\begin{figure}[h]
        \centering
        \includegraphics[width=1.0\linewidth]{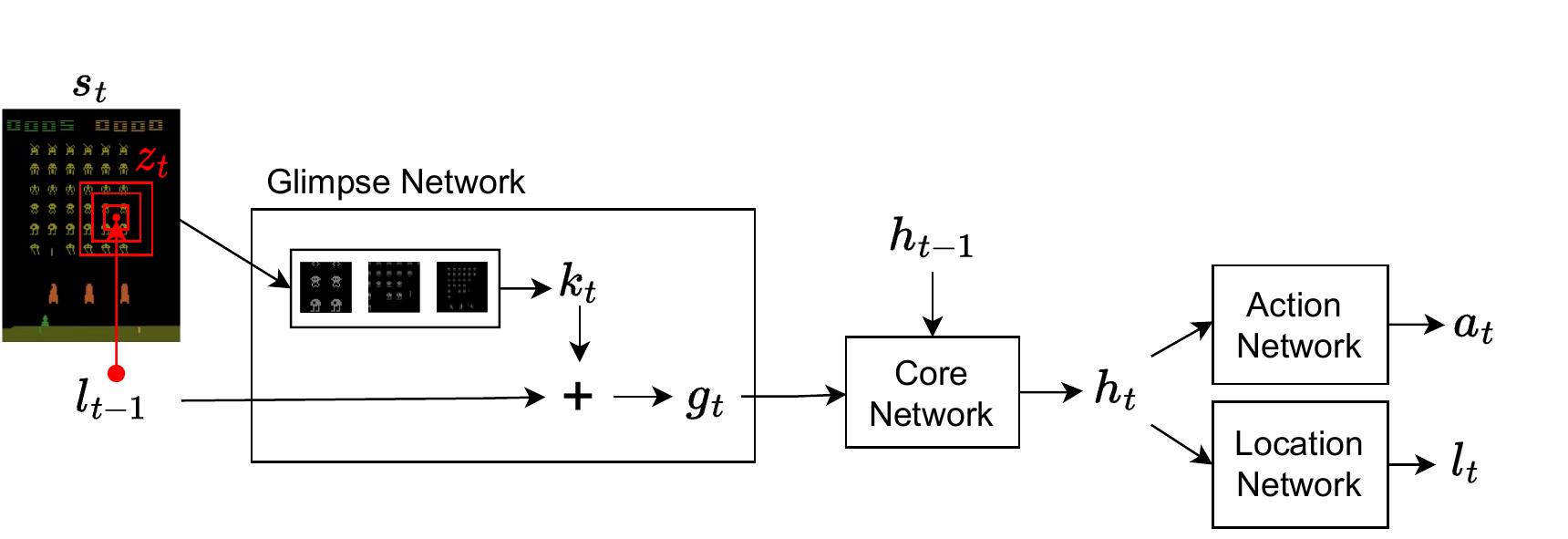}
        \caption{Architecture of RAM. Image adapted from the original paper of Mnih et al. \cite{mnih2014recurrent}}
        \label{fig:ram_overview}
    \end{figure}

The architecture of RAM \cite{mnih2014recurrent} was the basis for our proposal, as a result, we present it in Figure \ref{fig:ram_overview} and will briefly describe it next. 

Every timestep, RAM receives the game frame $s_{t}$ and a set of coordinates $l_{t-1}$. Its Glimpse Network takes a glimpse $z_{t}$ centered at $l_{t-1}$ and extracts features, not only from $z_{t}$, which are represented in Figure \ref{fig:ram_overview} by $k_{t}$, but also from $l_{t-1}$. Using fully connected layers the two features are merged into the vector $g_{t}$. Next, this vector is fed to an LSTM, the Core Network, which stores all the previous information from the glimpses and the coordinates chosen, building its internal memory $h_{t}$. After that, $h_{t}$ is the input for two other networks, the Action Network and the Location Network, which are also fully connected layers that select the next action $a_{t}$ and coordinates $l_{t}$, respectively. Since the Location Network was non-differentiable, it was trained using the policy gradient method REINFORCE, while the other components used backpropagation. 

After the presentation of RAM, some papers proposing improvements to the image classification capabilities of the model were written, such as the publications of Ba et al. \cite{ba2014multiple} and Zuur \cite{zuur2019deep}. After the presentation of our architecture, we will describe which refinements we took into consideration and how we have adapted them to our problem. 

    \begin{figure}[h]
        \centering
        \includegraphics[width=0.95\linewidth]{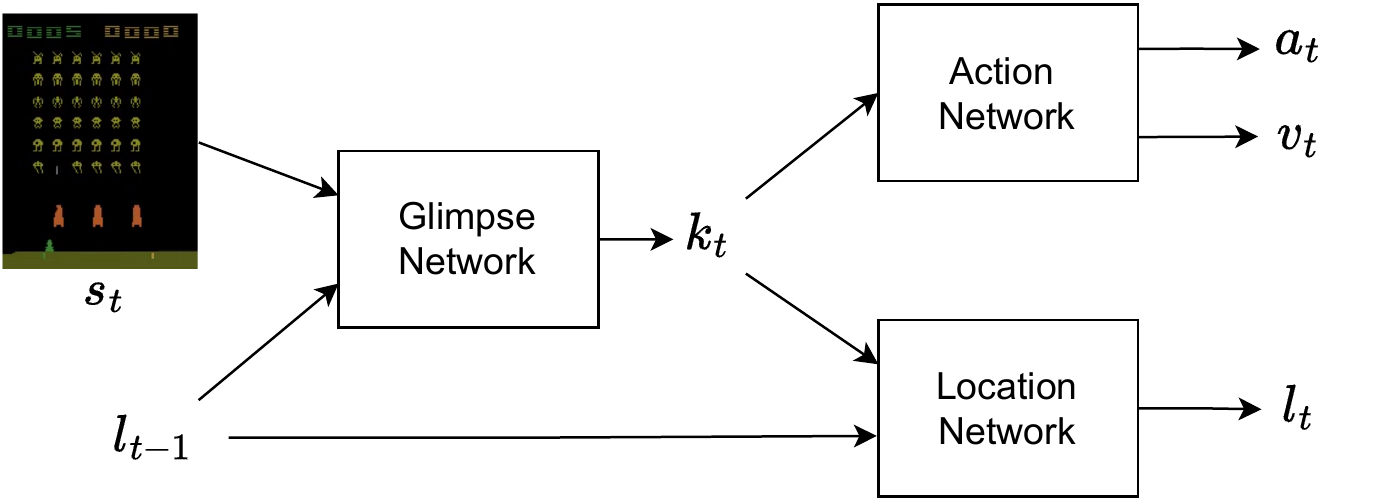}
        \caption{Overview of the architecture of the Glimpse-Based Actor-Critic}
        \label{fig:model_overview}
    \end{figure}

Figure \ref{fig:model_overview} presents an overview of the GBAC architecture. We start by receiving a game frame $s_{t}$ and the set of coordinates $l_{t-1}$ our agent chose at the end of the previous timestep. The Glimpse Network saves into $k_{t}$ the features extracted just from the glimpse taken from $s_{t}$ and centered at $l_{t-1}$. After that, the vector $k_{t}$ is used as the input of the Action Network. This network outputs not only the action $a_{t}$ the agent will take next, but also an estimate $v_{t}$ of the value function. $k_{t}$ is also used by the Location Network, which merges it with the features extracted from the location $l_{t-1}$ to select the next location coordinates $l_{t}$. In opposition to RAM, instead of doing this merging for the input of both networks, we only do it for the Location Network.

We now present the key refinements of RAM that we took into consideration to address our problem. To start, we followed Ba et al. \cite{ba2014multiple} and added a CNN to the Glimpse Network (represented in Figure \ref{fig:glimpse_network}) because it was necessary for the model to extract better features from the video frames. As one of the experiments done by Zuur \cite{zuur2019deep} suggested, we also found that it was beneficial to provide separate inputs for the Action and Location Networks (Figure \ref{fig:model_overview}). Therefore, the configuration which performed best was the one that fed to the Action Network only the features ($k_{t}$) extracted from the glimpse, and to the Location Network the vector resultant from the merge between $k_{t}$ and the features extracted from $l_{t-1}$. Still with this idea of separating the processes of choosing the next action and the next glimpse coordinates, we added an extra LSTM, making the Action Network and the Location Network use their separate LSTM (Figures \ref{fig:action_network} and \ref{fig:location_network}). This separation destroyed the need for having an explicit Core Network like in RAM, avoiding the mix of information that is necessary for each task and allowing us to fine-tune the parameters for each LSTM. The last change we made was the selection of PPO instead of REINFORCE to train the model since it achieves better results in harder environments \cite{zuur2019deep} and we found it simpler to train. This change meant that in each timestep, our Action Network also needed to make a prediction $v_{t}$ about the quality of performing the action $a_{t}$ in the current state of the environment.

\subsubsection{Glimpse Network}

\begin{figure}[h]
        \centering
        \includegraphics[width=0.95\linewidth]{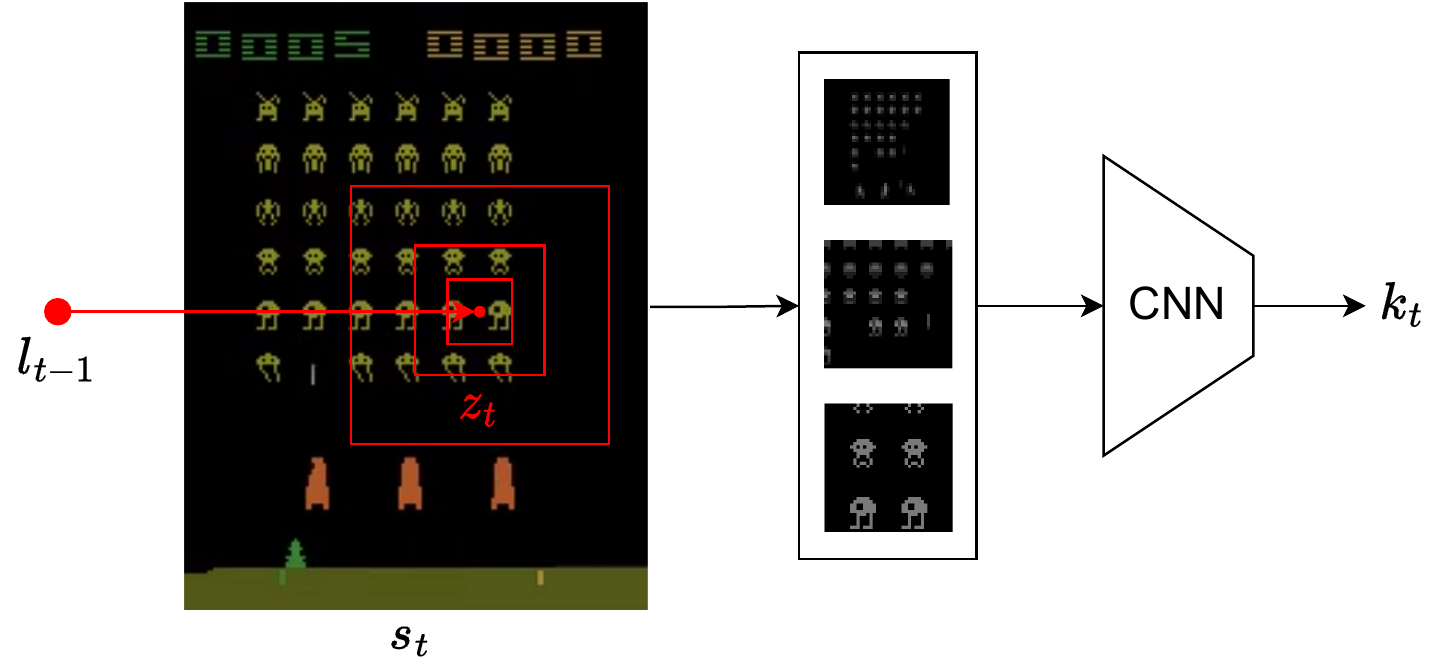}
        \caption{Detailed diagram of the Glimpse Network of GBAC}
        \label{fig:glimpse_network}
    \end{figure}

The Glimpse Network, represented in Figure \ref{fig:glimpse_network}, is the module responsible for extracting from the game frame, the region the agent chose to focus its attention. With the image coordinates $l_{t-1}$ chosen in the previous timestep, this network extracts an observation $z_{t}$ from $s_{t}$, which is called a \textit{glimpse}. This glimpse can have multiple patches, each having double the size of the previous. For example, Figure \ref{fig:glimpse_network} presents a glimpse with three patches. However, to simulate peripheral vision, all the larger patches are downscaled to the dimension of the smallest. That smallest patch will have the highest resolution, making it the focal point. Regarding the other patches, the larger they are, the further away they are from the focus point, so the lower their resolution is. When compared to the original image, this process results in a vector with fewer pixels. 

After rescaling, the resultant vector passes through a set of convolutional layers and a fully connected layer to extract a vector of features $k_{t}$. The number of convolutional layers and their respective kernel sizes and stride are changed depending on the size of the glimpse.

During this work, we assumed that the glimpses are always squares. If the coordinates of $l_{t}$ make a patch catch pixels that are out of the bounds of $s_{t}$, that patch will be moved in order to fit inside the frame. This mechanism proved to achieve better results than simply filling the pixels out of bounds with a value.
    
\subsubsection{Action Network}

\begin{figure}[h]
        \centering
        \includegraphics[width=0.95\linewidth]{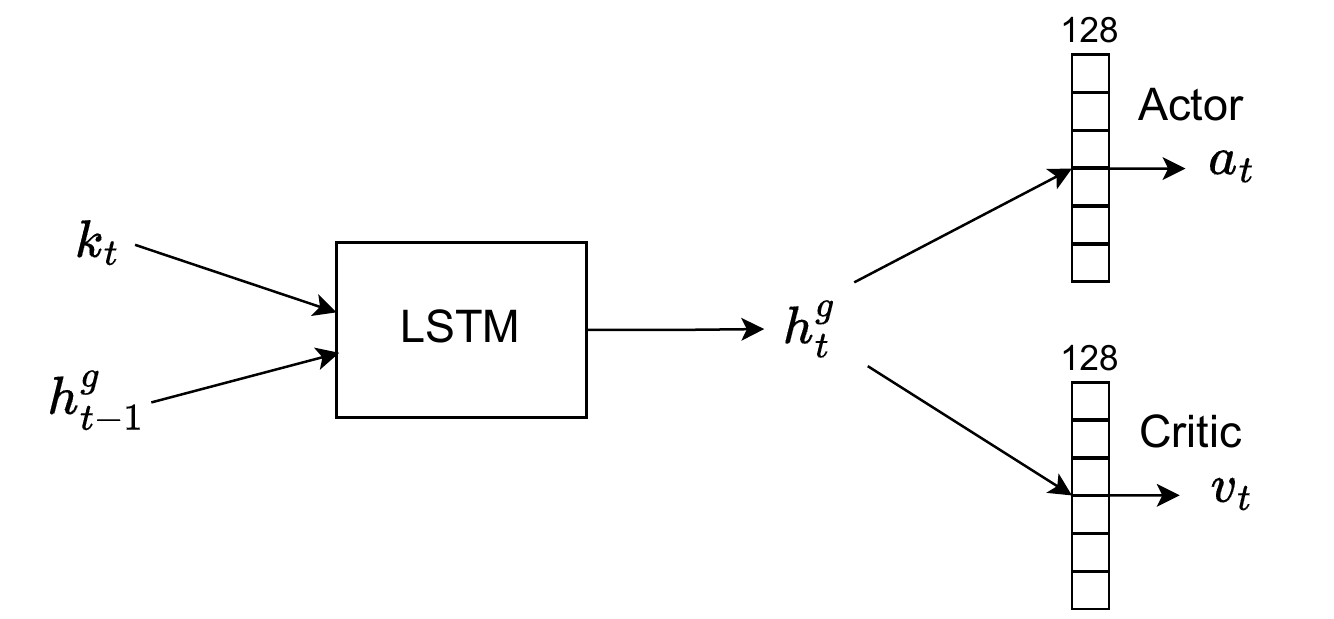}
        \caption{Detailed diagram of the Action Network of GBAC}
        \label{fig:action_network}
    \end{figure}

The Action Network, depicted in Figure \ref{fig:action_network}, is responsible to choose the game action $a_{t}$ the agent should perform in each timestep. Since we chose an actor-critic algorithm to train GBAC, the Action Network also estimates the value function, which helps the agent to understand if it is performing well or not.

In this network, its LSTM saves the information $z_{t}$ extracted previously from the glimpse and combines it with its hidden memory $h_{t-1}^{g}$, which stores all the information gathered in earlier timesteps to build an internal representation of the game environment. The new hidden state $h_{t}^{g}$ is fed to two fully connected layers, the Actor and the Critic, that output the action $a_{t}$ and the value $v_{t}$, respectively. 

\subsubsection{Location Network}

\begin{figure}[h]
        \centering
        \includegraphics[width=0.99\linewidth]{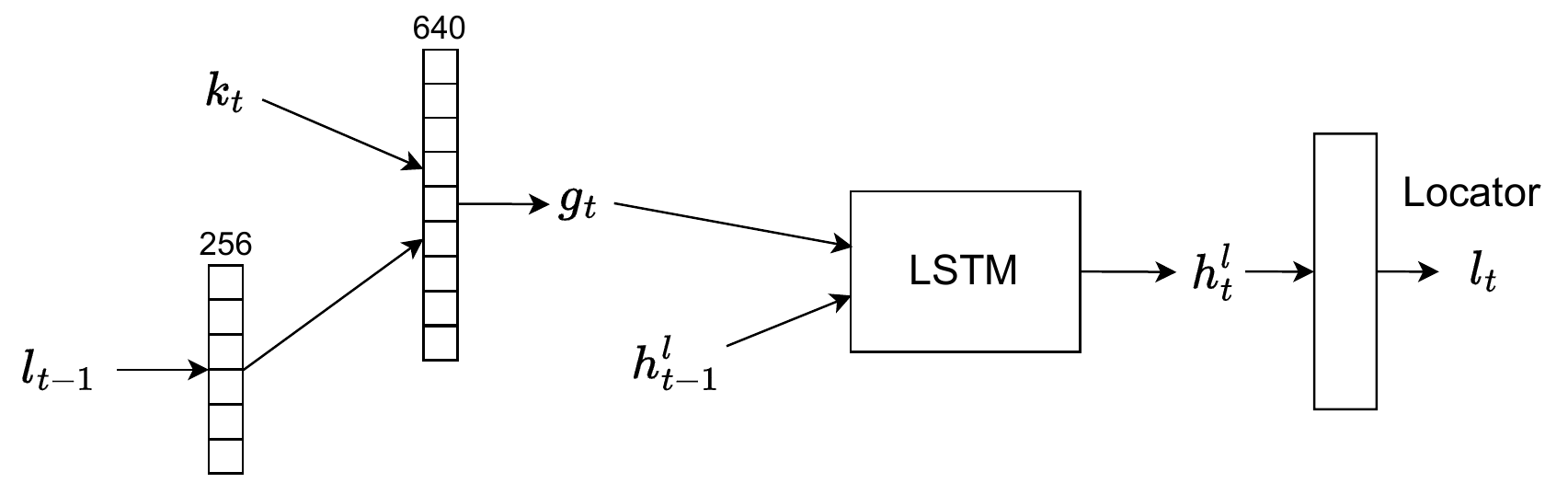}
        \caption{Detailed diagram of the Location Network of GBAC}
        \label{fig:location_network}
    \end{figure}

The Location Network, illustrated in Figure \ref{fig:location_network}, is the module behind the hard attention mechanism and is responsible to choose the image coordinates $l_{t}$ where the agent should look in the next timestep. Those coordinates are sampled from a truncated normal distribution whose mean is given by this network, being the standard deviation a fixed value. 

In order to choose the mean value, the Location Network has a neural network that extracts features from the coordinates $l_{t-1}$, merging them with the features $k_{t}$. The resultant vector $g_{t}$ is then used to update the internal state of an LSTM. Its internal state $h_{t}^{l}$ is fed to the Locator, which is a neural network with two fully connected layers and ReLU and Tanh activation functions, respectively. The Locator chooses the mean value used in the truncated normal distribution from which the next set of coordinates $l_{t}$ is extracted.

Since this model only uses a small portion of an image, its complexity is not dependent on the size of the input image but rather on the size of its glimpses. This can be an advantage if the model has to handle large images.

\subsection{Training}

The training process of our model is very similar to the one presented by Schulman et al. in the PPO paper \cite{schulman2017proximal}. We follow the suggestions they proposed, and the only difference is that we have two policy losses, instead of just one. Like PPO, our model alternates between interacting with the environment to get new information and updating both its policies with that new data. 

When exploring the environment, in each timestep, the model saves the action and coordinates of the glimpse it chose, the value function estimated by the critic, the reward received from the game, and a flag that indicates if the current episode has finished.

After a predetermined number of timesteps, the model uses the collected data to update its policies. For both policies, we use the "surrogate" objective already presented in Section \ref{sec:back_rl}:

\begin{equation}
    L^{CLIP_{\pi}}(\theta) = \hat{\mathbb{E}}_{t} \Big[ \min(r_{t}^{\pi}(\theta)\hat{A}_{t}, \mathrm{clip}(r_{t}^{\pi}(\theta), 1 - \epsilon, 1 + \epsilon)\hat{A}_{t}) \Big]
\end{equation}

where $\epsilon$ is a hyperparameter and the advantage function $\hat{A}_{t}$ is computed recurring to Generalized Advantage Estimation \cite{schulman2015high}.

Besides the two policy losses, the objective has two more terms: an entropy bonus $B$ that promotes the exploration of the environment, and the squared-error loss of the value function, $L_{t}^{VF}$. With these additions, the objective's formula is the following:

\begin{equation}
    L_{t}(\theta) = \hat{\mathbb{E}}_{t} \Big[ L_{t}^{CLIP_{\mathrm{a}}}(\theta) + L_{t}^{CLIP_{\mathrm{g}}}(\theta) - \alpha L_{t}^{VF}(\theta) + \beta B[\pi_{a}](s_{t})  \Big] 
\end{equation}

where $\alpha$ and $\beta$ are coefficients. Note that the entropy bonus is only calculated for the action policy, not for both policies. Adding the same bonus for the location policy did not seem to improve the results, thus we decided to keep the objective simpler. 

\section{Experimental Evaluation}
    In this chapter, we present all the experiments that enabled us to test our model, establishing which base PPO algorithm is the fairest choice to compare our model with, and discovering which size of the glimpses gives better results. In the end, we have all the information necessary to answer the questions in Section \ref{sec:intro}. Our code is available on \href{https://github.com/GonQue/gbac}{GitHub}.


\subsection{Evaluation Process}

\begin{table*}[]
\centering
\begin{tabular}{ll|rr|rr|cc|}
\cline{3-8}
 &
   &
  \multicolumn{2}{c|}{PongNoFrameskip-v4} &
  \multicolumn{2}{c|}{SpaceInvadersNoFrameskip-v4} &
  \multicolumn{2}{c|}{CarRacing-v0} \\ \hline
\multicolumn{1}{|c|}{\textbf{Model}} &
  \multicolumn{1}{c|}{\textbf{Full Img.}} &
  \multicolumn{1}{c|}{\textbf{Max. Train Avg.}} &
  \multicolumn{1}{c|}{\textbf{Test Avg.}} &
  \multicolumn{1}{c|}{\textbf{Max. Train Avg.}} &
  \multicolumn{1}{c|}{\textbf{Test Avg.}} &
  \multicolumn{1}{c|}{\textbf{Max. Train Avg.}} &
  \textbf{Test Avg.} \\ \hline
\multicolumn{1}{|l|}{PPO} &
  No &
  \multicolumn{1}{r|}{21.00 $\pm$ 0.01} &
  20.98 $\pm$ 0.02 &
  \multicolumn{1}{r|}{2090.00 $\pm$ 254.16} &
  2013.07 $\pm$ 244.56 &
  \multicolumn{1}{c|}{--} &
  - \\ \hline
\multicolumn{1}{|l|}{PPO} &
  Yes &
  \multicolumn{1}{r|}{20.91 $\pm$ 0.11} &
  20.83 $\pm$ 0.13 &
  \multicolumn{1}{r|}{2261.90 $\pm$ 295.87} &
  2221.62 $\pm$ 201.31 &
  \multicolumn{1}{r|}{867.16 $\pm$ 6.64} &
  \multicolumn{1}{r|}{824.31 $\pm$ 8.04} \\ \hline
\multicolumn{1}{|l|}{PPO + LSTM} &
  No &
  \multicolumn{1}{r|}{20.11 $\pm$ 0.25} &
  20.00 $\pm$ 0.31 &
  \multicolumn{1}{r|}{1182.57 $\pm$ 259.03} &
  1077.63 $\pm$ 245.82 &
  \multicolumn{1}{c|}{-} &
  - \\ \hline
\multicolumn{1}{|l|}{PPO + LSTM} &
  Yes &
  \multicolumn{1}{r|}{20.03 $\pm$ 0.23} &
  19.85 $\pm$ 0.39 &
  \multicolumn{1}{r|}{900.20 $\pm$ 79.16} &
  812.58 $\pm$ 111.51 &
  \multicolumn{1}{r|}{783.62 $\pm$ 11.58} &
  \multicolumn{1}{r|}{659.73 $\pm$ 24.42} \\ \hline
\end{tabular}
\caption{Comparison between the training and testing performance of PPO and PPO+LSTM, using a resized frame (84x84) of the input and also the full game frame (210x160)}
\label{tab:base_models}
\end{table*}

\subsubsection{Game Environments} \label{sec:game_envs}
In order to measure the performance of GBAC and compare it against the state-of-the-art RL agents, we selected three game environments with different characteristics. The first two are both games from the Atari 2600 and have a discrete action space, while the third, is CarRacing \cite{carRacing2016} from OpenAI Gym \cite{openAI2016} and has a continuous action space. We tried to select three games that were not the easiest ones available and that required the agents to learn different sets of skills, such that we could see how they were capable to adapt to each type of task.

In order to obtain the best performance in the Atari games, we used the same environment modifications proposed by Mnih et al. \cite{mnih2015human} and Machado et al. \cite{machado2018revisiting}, thus leading to better results for these games. Those modifications are well accepted and used in the literature. For example, we resize the original frame from 210x160 to 84x84 pixels, clip the rewards, and scale the pixel values to [0, 1].

\subsubsection{Comparison Models}

We decided to compare GBAC with three different versions of PPO. The first one is the original PPO agent presented by Schulman et al. \cite{schulman2017proximal} because it was used as the base for our model. However, since our implementation uses LSTMs in its architecture and the version of PPO with an LSTM is also quite common in the RL literature, we decided to choose the PPO+LSTM agent for comparison as well. The third version of PPO we used is a modification of the PPO+LSTM algorithm where the restriction of only using a small portion of the game frame was imposed. But, different from GBAC, the coordinates where the agent looks are chosen completely randomly. This allows us to test if the perception mechanism implemented in our model is better than one that makes its choices randomly. 

The PPO implementation used in this work was not the original one provided by Schulman et al. \cite{schulman2017proximal} in OpenAI Baselines \cite{baselines}, but rather a revised implementation presented by Shengyi et al. \cite{shengyi2022the37implementation} that closely follows the performance of the original. This implementation provides many versions of PPO including one with an LSTM. To make the PPO agents capable of playing the selected games, we added to their architecture a CNN with the same layout as the one presented in the DQN paper \cite{mnih2015human}. 

By comparing GBAC with the first two versions of PPO, we can discover if it is possible to achieve state-of-the-art performance playing video games, despite having a limited (but active) perception of the environment, which was our first question.


\subsubsection{Evaluation Metrics}
Evaluating RL models is never a straightforward task due to their high variance. For this reason, during training and testing, we average the episodic return each agent achieved over the last 100 episodes.

In training, Pong, SpaceInvaders, and CarRacing learned during 15 million, 20 million, and 5 million timesteps, respectively. Then, the agent that achieved the best performance over the last 100 episodes is tested for another 100 episodes. For each configuration, the results are always the average over three runs and their respective standard deviations are also presented.

Seeding is another aspect that can have an impact on the performance of the agent. Certain seeds can make the agent perform significantly better or worse. Thus, in each run, the seed used in the environment is chosen arbitrarily.

Regarding the policy that chooses the locations of the glimpses, in order to understand if our model learns a behavior that resembles the human vision, we analyze the evolution of the policy throughout the training phase and present a visual representation of the results. This is the information we need to answer our second question.

Now that the entire evaluation process is detailed, we present, in the next sections, the results we obtained.

\subsection{Base Models Selection} 

In contrast to one of the Atari optimizations proposed by Mnih et al. \cite{mnih2015human} and Machado et al. \cite{machado2018revisiting}, our model does not perform better when the original image is resized. In fact, it never learned a way to beat its adversary in Pong when a glimpse smaller than the resized size of 84x84 was used. This meant that with that setup, we could not take advantage of the glimpse's structure because the model needed the entire image to perform well, which does not follow the restriction of our problem. An explanation for this result can be the fact that since the glimpses taken by GBAC have multiple patches that end up being resized to a lower resolution, and the input frame fed to the model was also resized, the loss in information might be so much that our agent is not capable of solving the game. 

Therefore, to make the comparisons fair, we decided that the base PPO models should also use the entire image as input. In order to discover how much this decision could impact the performance of the base agents when playing the two Atari games, we studied the difference in performance between using the original 210x160 image and the resized 84x84 input. In CarRacing, since the original image is just 96x96, we decided not to compare the base models with a resized version of the input.
In addition, since GBAC uses LSTMs, we compared PPO with PPO+LSTM to find out if they perform any differently.

In Table \ref{tab:base_models}, we show the training and testing performances that PPO and PPO+LSTM had when using either the full image frame or the resized input. The results shown do not allow us to conclude with absolute certainty that one way is better than the other. In Pong, there are not any significant differences in performance either in regular PPO or in PPO+LSTM. In SpaceInvaders, for the regular PPO, the agent that uses the full image achieves average returns that are around 200 points better than the agent that resizes the input. This result indicates that, for this game, some useful information is lost during the resizing of the image. However, for the model that uses PPO+LSTM, the opposite is verified. Since the size of the LSTMs was the same in both cases, this suggests that for the full image, the LSTM needed to be bigger because it could extract better data from fewer pixels.

From the same table, when comparing the PPO agent against PPO+LSTM for the same type of input, we can see that the use of an LSTM deteriorates the performance of the agent. 
Having in mind that in the PPO game, a match ends when a player reaches 21 points and the reward of the agent is the difference between the points scored and scored against, the difference between the two models is marginal. It only means that, on average, the PPO+LSTM agent let the opponent score one point, while PPO did not. In SpaceInvaders, the performance drops by half, which is a significant reduction. In CarRacing, there is also a drop in performance, even though it is less accentuated than in SpaceInvaders. The major difference between the PPO and PPO+LSTM models is that the former uses a stack of four frames as input, while the latter receives just one frame at a time, counting on its LSTM to discover and store the information that is useful to the agent. Therefore, in SpaceInvaders and in CarRacing, the LSTM is not able to store all the information needed, like the velocity and direction of the objects from the game, which would allow the agent to perform better.  

\smallskip

In short, from these results, we cannot conclude that using the resized frame is better than using the full frame, and since our model utilizes the full image, in order to make the comparisons fairer, with as many equal variables as possible, from here onwards, we will be referring to the version that uses an LSTM and the entire frame as input when mentioning the base model.

\subsection{GBAC Performance Analysis}

In this section, we study not only the impact that different sizes of glimpses and different numbers of patches have on the performance of our agent but also how the best configuration performs against the three versions of PPO. 

In our architecture, each glimpse can have one or more patches. Since we stipulated that each new patch has double the size of the previous, increasing the number of patches results in glimpses with a smaller focus region. This means that if we want glimpses with two patches, the largest possible size for the smallest patch is 80x80, with three patches it will be 40x40, and with four patches 20x20. The higher the number of patches, the larger the "peripheral vision" of our model. Nonetheless, this increase in information comes with the price of it not being as detailed as the portions of the image closer to the focal point.

With this in mind, besides their architectures, using glimpses with just one patch makes our model no different from the base PPO agents. Therefore, the results from the agents that use glimpses with one patch are just useful to compare the two architectures, and to discover how large the input image has to be, in order for the agent to maintain its performance. Therefore, the results that are relevant to understand the performance of our model are the ones given by configurations that use more than one patch.

\subsubsection{Pong}



In relation to the Atari game Pong, Table \ref{tab:comparison_models_pong} presents the maximum training average and the testing performance of the three PPO versions, as well as the best glimpse size for each number of patches of our agent. Since the returns for this game are bounded between -21 and +21, when analyzing the performance of all the agents, we can see that one of three scenarios occurred: the agent learned how to beat its adversary and ended up with scores close to +20; the agent did not manage to learn anything useful, resulting in scores around -19; or the timesteps were not enough for the agent to learn a good policy so it achieved a score between the previous two.

\begin{table}[ht]
\centering
\begin{tabular}{lc|rr|}
\cline{3-4}
                               & \multicolumn{1}{l|}{} & \multicolumn{2}{c|}{PongNoFrameskip-v4}                    \\ \hline
\multicolumn{1}{|c|}{\textbf{Model}} & \textbf{Glimpse} & \multicolumn{1}{c|}{\textbf{Max. Train Avg.}} & \multicolumn{1}{c|}{\textbf{Test Avg.}} \\ \hline
\multicolumn{1}{|l|}{PPO}      & -                     & \multicolumn{1}{r|}{20.91 $\pm$ 0.11}  & 20.83 $\pm$ 0.13  \\ \hline
\multicolumn{1}{|l|}{PPO+LSTM} & -                     & \multicolumn{1}{r|}{20.03 $\pm$ 0.23}  & 19.85 $\pm$ 0.39  \\ \hline
\multicolumn{1}{|l|}{GBAC}     & 1p, 160x160           & \multicolumn{1}{r|}{7.61 $\pm$ 10.42}  & 6.95 $\pm$ 10.89  \\ \hline
\multicolumn{1}{|l|}{GBAC}     & 2p, 80x80             & \multicolumn{1}{r|}{7.15 $\pm$ 20.80}  & 6.64 $\pm$ 21.14  \\ \hline
\multicolumn{1}{|l|}{GBAC}     & 3p, 40x40             & \multicolumn{1}{r|}{20.06 $\pm$ 0.44}  & 19.82 $\pm$ 0.88  \\ \hline
\multicolumn{1}{|l|}{GBAC}     & 4p, 20x20             & \multicolumn{1}{r|}{-14.86 $\pm$ 5.39} & -15.77 $\pm$ 4.82 \\ \hline
\multicolumn{1}{|l|}{PPO Random} & 3p, 40x40             & \multicolumn{1}{r|}{-19.87 $\pm$ 0.05} & -20.16 $\pm$ 0.11 \\ \hline
\end{tabular}
\caption{Training and testing performance in Pong}
\label{tab:comparison_models_pong}
\end{table}

Regarding the glimpses with one patch, our agent did not manage to achieve scores near +20 with the larger glimpse consistently because, in two of the three runs, the agent was still improving its performance when the timesteps finished. Therefore in this game, our agent was not capable of matching the performances of PPO. With two patches, the results were slightly better than the previous because the standard deviation is much higher. The model achieved a score of around +19 in two of the three runs. Using three patches, GBAC was capable of matching the performance of the PPO+LSTM agent, consistently beating its opponent by 21-1. The difference to the regular PPO agent, only means that our agents let the opponent score a point, while the other did not. After seeing the results until this point, we might think that the performance keeps improving while we increase the number of patches of each glimpse. However, this is not the case when we look at the scores achieved using four patches. The size of the patches was so small that our model was not capable of achieving a performance of +20 in, at least, one of the three runs, which was true in any of the previous results.

Regarding the PPO agent that took glimpses at random locations, we see that it performed very poorly, not being able to learn an optimal policy to play Pong. Therefore, we can say that, in this game, choosing good glimpse locations matters.

\subsubsection{SpaceInvaders}

Table \ref{tab:comparison_models_spaceinvaders} shows the training and testing results of all the agents for the SpaceInvaders game.

\begin{table}[ht]
\centering
\begin{tabular}{lc|rr|}
\cline{3-4}
                                 & \multicolumn{1}{l|}{} & \multicolumn{2}{c|}{SpaceInvadersNoFrameskip-v4}                 \\ \hline
\multicolumn{1}{|c|}{\textbf{Model}} & \textbf{Glimpse} & \multicolumn{1}{c|}{\textbf{Max. Train Avg.}} & \multicolumn{1}{c|}{\textbf{Test Avg.}} \\ \hline
\multicolumn{1}{|l|}{PPO}        & -                     & \multicolumn{1}{r|}{2261.90 $\pm$ 295.87} & 2221.62 $\pm$ 201.31 \\ \hline
\multicolumn{1}{|l|}{PPO+LSTM}   & -                     & \multicolumn{1}{r|}{900.20 $\pm$ 79.16}   & 812.58 $\pm$ 111.51  \\ \hline
\multicolumn{1}{|l|}{GBAC}       & 1p, 160x160           & \multicolumn{1}{r|}{741.45 $\pm$ 392.54}  & 607.42 $\pm$ 287.84  \\ \hline
\multicolumn{1}{|l|}{GBAC}       & 2p, 80x80             & \multicolumn{1}{r|}{444.18 $\pm$ 40.78}   & 341.47 $\pm$ 25.71   \\ \hline
\multicolumn{1}{|l|}{GBAC}       & 3p, 40x40             & \multicolumn{1}{r|}{596.50 $\pm$ 182.35}  & 544.43 $\pm$ 166.79  \\ \hline
\multicolumn{1}{|l|}{GBAC}       & 4p, 20x20             & \multicolumn{1}{r|}{439.72 $\pm$ 26.27}   & 378.00 $\pm$ 28.70   \\ \hline
\multicolumn{1}{|l|}{PPO Random} & 3p, 40x40             & \multicolumn{1}{r|}{516.62 $\pm$ 60.59}   & 467.38 $\pm$ 50.53   \\ \hline
\end{tabular}
\caption{Training and testing performance in SpaceInvaders}
\label{tab:comparison_models_spaceinvaders}
\end{table}

Starting with the results of our model for one patch, we found that this time, the average score of GBAC was closer to the one achieved by the PPO+LSTM agent. However, when considering the results for the glimpses with two patches, the performance dropped almost by half. With three patches, GBAC achieves the best performance when considering the use of more than one patch. Nonetheless, the model is not able to match the performance achieved when it used the bigger glimpse with just one patch. With four patches, we have the same decrease in performance, already seen in Pong. However, this time it was not as severe and slightly beat the results from two patches while having the smallest standard deviation in both training and testing of any of the experiments.

In this game, our model was not capable of matching the performance of the PPO+LSTM agent, however, we still consider it an interesting result, considering the viewing restrictions of our problem. While processing 86\% fewer pixels than PPO+LSTM (4.800 vs. 33.600) in each timestep, GBAC only had a performance drop of 33\%.

In this game, the random agent has a performance that is on par with our model, which possibly means that in SpaceInvaders the location of a glimpse is not that important. One possible reason behind this proximity could be the fact that, in this game, GBAC can pay attention to a lot more things that can improve the return received from the environment. As opposed to Pong, where our agent only had three objects (two paddles and one ball) to keep track of. 

    \begin{figure*}[ht]
    	\centering
        \begin{subfigure}{0.24\textwidth}
            \centering
            \includegraphics[width=0.60\linewidth]{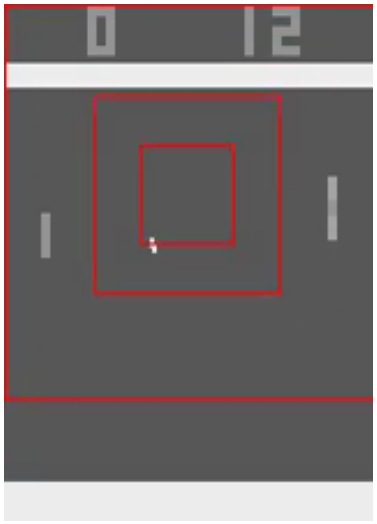}
            \caption{Pong 40x40 Glimpse}
            \label{fig:pong_glimpse}
            \Description{Glimpse of 40x40 with three patches in Pong} 
        \end{subfigure}
        \hfill
        \begin{subfigure}{0.24\textwidth}
            \centering
            \includegraphics[width=0.85\linewidth]{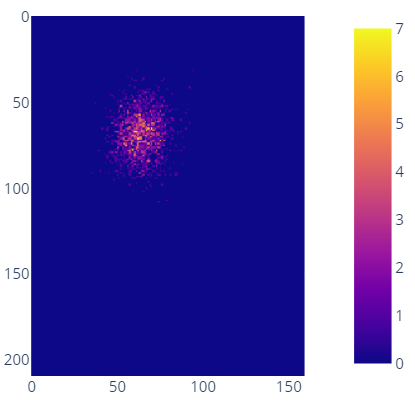}
            \caption{Pong Heatmap}
            \label{fig:pong_heatmap}
            \Description{In the case of Pong, the locations chosen by the model have the shape of a normal distribution in two dimensions and are positioned slightly upwards and leftwards from the center of the image frame.}
        \end{subfigure}
        \hfill
        \begin{subfigure}{0.24\textwidth}
            \centering
            \includegraphics[width=0.60\linewidth]{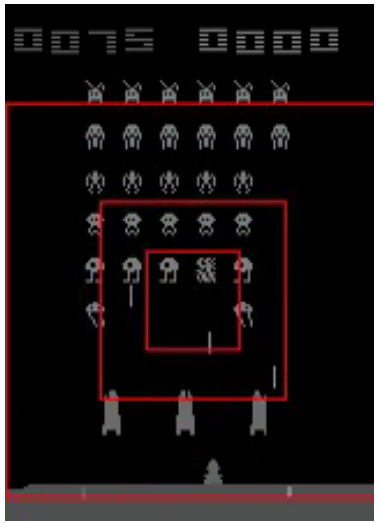}
            \caption{SpaceInvaders 40x40 Glimpse}
            \label{fig:spaceinvaders_glimpse}
            \Description{Glimpse of 40x40 with three patches in SpaceInvaders}
        \end{subfigure}
        \hfill
        \begin{subfigure}{0.24\textwidth}
            \centering
            \includegraphics[width=0.89\linewidth]{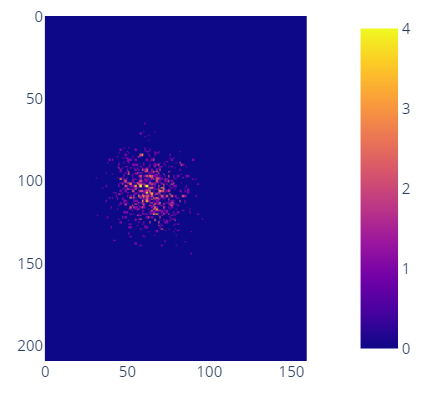}
            \caption{SpaceInvaders heatmap}
            \label{fig:spaceinvaders_heatmap}
            \Description{In the case of SpaceInvaders, the locations have the same distribution shape, although, for this game, they are more sparse and just positioned slightly leftwards from the center.}
        \end{subfigure}
        \caption{Frames with glimpses of 3 patches and heatmaps representing the number of times during an episode that the agent chose a specific set of image coordinates to be the center of the glimpse, for the Pong and SpaceInvaders games respectively}
    	\label{fig:atari_images}
        \Description{Examples of a glimpse in Pong and SpaceInvaders and heatmaps representing the number of times during an episode that the agent chose a specific set of image coordinates to be the center of the glimpse.} 
    \end{figure*}

\subsubsection{CarRacing}

Lastly, Table \ref{tab:comparison_models_carracing} shows the performance that GBAC and the other agents achieved during training and testing, when playing the CarRacing game, which has a game frame of size 96x96.

\begin{table}[ht]
\centering
\begin{tabular}{lc|rr|}
\cline{3-4}
                                 & \multicolumn{1}{l|}{} & \multicolumn{2}{c|}{CarRacing-v0}                             \\ \hline
\multicolumn{1}{|c|}{\textbf{Model}} & \textbf{Glimpse} & \multicolumn{1}{c|}{\textbf{Max. Train Avg.}} & \multicolumn{1}{c|}{\textbf{Test Avg.}} \\ \hline
\multicolumn{1}{|l|}{PPO}        & -                     & \multicolumn{1}{r|}{867.16 $\pm$ 6.64}   & 824.31 $\pm$ 8.04  \\ \hline
\multicolumn{1}{|l|}{PPO+LSTM}   & -                     & \multicolumn{1}{r|}{783.62 $\pm$ 11.58}  & 659.73 $\pm$ 24.42 \\ \hline
\multicolumn{1}{|l|}{GBAC}       & 1p, 96x96             & \multicolumn{1}{r|}{815.12 $\pm$ 5.75}   & 660.02 $\pm$ 69.38 \\ \hline
\multicolumn{1}{|l|}{GBAC}       & 2p, 40x40             & \multicolumn{1}{r|}{694.50 $\pm$ 107.94} & 641.11 $\pm$ 57.42 \\ \hline
\multicolumn{1}{|l|}{GBAC}       & 3p, 20x20             & \multicolumn{1}{r|}{676.82 $\pm$ 84.89}  & 564.00 $\pm$ 56.42 \\ \hline
\multicolumn{1}{|l|}{PPO Random} & 2p, 40x40             & \multicolumn{1}{r|}{622.41 $\pm$ 17.89}  & 589.87 $\pm$ 13.19 \\ \hline
\end{tabular}
\caption{Training and testing performance in CarRacing}
\label{tab:comparison_models_carracing}
\end{table}

Like in the other two games, here, the largest glimpse size is also the one that achieves the better result for a glimpse with one patch. In addition, this time our model was capable of matching the performance of the PPO+LSTM agent in testing and beating it during training. For two patches, the achieved results maintained the performance seen in testing when using one patch, even though the score in training decreased and had a much higher standard deviation. With three patches, which in CarRacing was the maximum number we tested, we start to see the performance drop slightly.

In this game, GBAC was capable of maintaining its performance across all number of patches, which is a good result since the scores closely match the PPO+LSTM agent.

Regarding the PPO agent with random glimpses, the performance difference is again not that far behind our model. Since the generated track occupies a good portion of the image, it may be easier for the random model to select good actions from every glimpse it receives.

From this study, we can conclude that even when using the entire frame (glimpses with one patch) the performance of our model is capable of matching the performance of PPO+LSTM, although not consistently. This shows our architecture could compete with PPO+LSTM if we did not impose our restrictions. Additionally, we discovered that the performance of GBAC does not increase linearly with the number of patches. It reaches a point where the information lost with the reduction of the glimpse size is more significant than the information gained with the addition of another patch. The optimal number of patches is three for the Atari games and two for CarRacing. Those numbers of patches proved to be the right balance between having a patch size that discarded the irrelevant information, allowing the model to just focus on the most important, and not being too small such that after rescaling the large patches, it was still possible to understand what was present in the "peripheral vision" of the agent. Finally, we saw that in most cases, the largest glimpse size possible for each number of patches is the one that produces the best results. In CarRacing, for two and three patches this was not the case, and the sizes 48x48 and 24x24, respectively, did not give the best results, even though they were very close. 

An important fact we should mention is that, since the complexity of our model is independent of the size of the input, we can achieve these results with a model that, in the Atari games, has 15\% fewer total training parameters ($\sim$1.7M vs. $\sim$2.0M) than the PPO+LSTM model, and almost the same number has PPO. In CarRacing, since the actions are continuous, the model is bigger, but it still has 10\% fewer parameters than PPO+LSTM ($\sim$2.2M vs. $\sim$2.5M) and only more 70k than PPO. If we use bigger environments having frames with many more pixels, this difference between the number of training parameters required will only keep increasing.  

\subsection{Glimpse Movements Analysis}

After discovering how GBAC performs against the other models, it is also important to understand how the location of its glimpses is evolving throughout training and which behavior the model finds outs to be the best. While making this analysis, we also compare the agent's decisions with the choices we would consider when playing the games.  

Regarding the regions that our best agents chose to look at during the episodes, we can see in Figure \ref{fig:pong_heatmap} and Figure \ref{fig:spaceinvaders_heatmap} that, for the Atari games, their distribution is relatively similar in both games. In SpaceInvaders, our agent keeps its focus near the center of the image, while in Pong, it chooses locations slightly upwards from the center. In general, the distribution of locations in both games has more choices closer to the center and becomes more sparse the further they are from it.

Having the focus point almost near the center of the image, as we can see in Figure \ref{fig:pong_heatmap}, means that, in Pong, the agent gets the location of each paddle from its "peripheral vision" (Figure \ref{fig:pong_glimpse}). We think that this choice is different from what a human would select to focus on in this particular game. We believe that a human would pay more attention to the position of its own paddle and the ball.

In relation to SpaceInvaders, in our opinion, the choices are much closer to what a human would do because the agent keeps its focus on the lower rows of enemies (Figure \ref{fig:spaceinvaders_glimpse}), which are the ones that need to be destroyed first.

    \begin{figure}[ht]
    	\centering
        \begin{subfigure}{0.23\textwidth}
            \centering
            \includegraphics[width=0.90\linewidth]{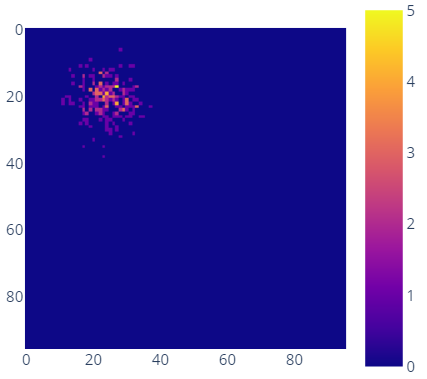}
            \caption{Beginning of training}
            \label{fig:carracing_heatmap_1}
            \Description{At the beginning of the training of CarRacing, the locations have the same distribution shape as the other two games, although, for in this particular case, it is even more upwards and leftwards than in Pong, being almost near the top-left corner of the game frame.}
        \end{subfigure}
        \hfill
        \begin{subfigure}{0.23\textwidth}
            \centering
            \includegraphics[width=0.90\linewidth]{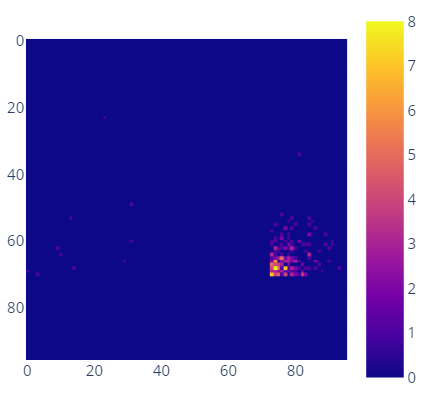}
            \caption{Middle of training}
            \label{fig:carracing_heatmap_3}
            \Description{At the middle of the training of CarRacing, the locations stop having a circular shape and start forming a squared shape distribution with the most points being concentrated in the bottom-left corner of the square. Now, that square distribution is almost in the bottom-right corner of the game frame instead of in the top-left as registered previously.}
        \end{subfigure}
    	\caption{Heatmaps representing the evolution of glimpse locations in the CarRacing game}
    	\label{fig:carracing_heatmaps}
    \end{figure}

Regarding Car Racing, our model presents quite unusual behavior during the training process. It starts similarly to the other two games with a circular normal distribution for the location coordinates (Figure \ref{fig:carracing_heatmap_1}). However, after a few training epochs, that distribution starts dispersing over multiple parts of the entire image, ending up with the region shown in Figure \ref{fig:carracing_heatmap_3}, creating some kind of borders, which constrain the choice of coordinates, and that do not correspond to the limits of the image. This last behavior is the one that is present in the best solution at the end of training.

In our opinion, even though in two of the three games, our model does not follow exactly what we think is the human vision behavior when playing those video games, we would need a more systematic way of analyzing the regions that we select to focus our attention, using, for example, an eye-tracking device, to better compare the agent's behavior in relation to humans.

\section{Conclusion}
    This work proposed a solution for the problem of an agent that has limited vision, and for that reason, besides deciding which action it has to take in the environment, it also has to choose which part of the environment it should look at. To solve this problem, we proposed GBAC, a model that combines a glimpse-based hard attention mechanism with a model-free RL algorithm.

We started by proving that, for some games like Pong and CarRacing, our model is already capable of achieving similar performance to the PPO version that more closely resembles our model, that is, the variant that also uses an LSTM. On other games like SpaceInvaders, a drop in performance is verified, with means, there still is room for improvement.
We finished concluding that our model does not necessarily choose the same regions of the image that we selected to look at when we played the video games ourselves. Only in SpaceInvaders we considered this was verified. In addition, in CarRacing, we are not able to explain the reason behind the unusual behavior of our model.

\begin{acks}
    This work was partially supported by Portuguese national funds through Fundação para a Ciência e a Tecnologia (FCT) under projects UIDB/50021/2020 (INESC-ID multi-annual funding), PTDC/CCI-COM/5060/2021 (RELEvaNT), PTDC/CCI-COM/7203/2020 (HOTSPOT). In addition, this research was partially supported by the Air Force Office of Scientific Research under award number FA9550-22-1-0475 and an EU Horizon 2020 project (TAILOR) under GA No. 952215.
\end{acks}

\bibliographystyle{ACM-Reference-Format}
\bibliography{references}

\newpage
\onecolumn
\appendix
    \section{Extended Results}

\begin{table}[ht]
\centering
\begin{tabular}{cc|rr|rr|}
\cline{3-6}
\multicolumn{1}{l}{} &
  \multicolumn{1}{l|}{} &
  \multicolumn{2}{c|}{PongNoFrameskip-v4} &
  \multicolumn{2}{c|}{SpaceInvadersNoFrameskip-v4} \\ \hline
\rowcolor[HTML]{AFAFAF} 
\multicolumn{1}{|c|}{\cellcolor[HTML]{AFAFAF}No. Patches} &
  Glimpse Size &
  \multicolumn{1}{c|}{\cellcolor[HTML]{AFAFAF}Max. Train Avg.} &
  \multicolumn{1}{c|}{\cellcolor[HTML]{AFAFAF}Test Avg.} &
  \multicolumn{1}{c|}{\cellcolor[HTML]{AFAFAF}Max. Train Avg.} &
  \multicolumn{1}{c|}{\cellcolor[HTML]{AFAFAF}Test Avg.} \\ \hline
\multicolumn{1}{|c|}{1} &
  160 &
  \multicolumn{1}{r|}{\cellcolor[HTML]{9AFF99}7.61 $\pm$ 10.42} &
  \cellcolor[HTML]{9AFF99}6.95 $\pm$ 10.89 &
  \multicolumn{1}{r|}{\cellcolor[HTML]{9AFF99}741.45 $\pm$ 392.54} &
  \cellcolor[HTML]{9AFF99}607.42 $\pm$ 287.84 \\ \hline
\multicolumn{1}{|c|}{1} &
  150 &
  \multicolumn{1}{r|}{-18.96 $\pm$ 1.67} &
  -19.32 $\pm$ 1.38 &
  \multicolumn{1}{r|}{384.43 $\pm$ 72.19} &
  338.45 $\pm$ 59.50 \\ \hline
\multicolumn{1}{|c|}{1} &
  140 &
  \multicolumn{1}{r|}{-18.93 $\pm$ 1.80} &
  -19.36 $\pm$ 1.47 &
  \multicolumn{1}{r|}{386.50 $\pm$ 75.94} &
  347.08 $\pm$ 99.27 \\ \hline
\multicolumn{1}{|c|}{1} &
  130 &
  \multicolumn{1}{r|}{-19.92 $\pm$ 0.04} &
  -20.19 $\pm$ 0.03 &
  \multicolumn{1}{r|}{330.76 $\pm$ 77.16} &
  265.00 $\pm$ 53.71 \\ \hline
\multicolumn{1}{|c|}{1} &
  120 &
  \multicolumn{1}{r|}{-19.86 $\pm$ 0.04} &
  -20.06 $\pm$ 0.05 &
  \multicolumn{1}{r|}{343.98 $\pm$ 137.35} &
  273.00 $\pm$ 113.99 \\ \hline
\multicolumn{1}{|c|}{1} &
  110 &
  \multicolumn{1}{r|}{-19.89 $\pm$ 0.11} &
  -20.22 $\pm$ 0.08 &
  \multicolumn{1}{r|}{381.66 $\pm$ 85.91} &
  340.12 $\pm$ 77.90 \\ \hline
\multicolumn{1}{|c|}{1} &
  100 &
  \multicolumn{1}{r|}{-19.89 $\pm$ 0.05} &
  -20.24 $\pm$ 0.11 &
  \multicolumn{1}{r|}{321.91 $\pm$ 51.51} &
  256.78 $\pm$ 77.16 \\ \hline
\multicolumn{1}{|c|}{1} &
  90 &
  \multicolumn{1}{r|}{-19.65 $\pm$ 0.42} &
  -19.90 $\pm$ 0.36 &
  \multicolumn{1}{r|}{285.76 $\pm$ 61.01} &
  214.35 $\pm$ 23.69 \\ \hline
\multicolumn{1}{|c|}{1} &
  80 &
  \multicolumn{1}{r|}{-19.90 $\pm$ 0.01} &
  -20.28 $\pm$ 0.05 &
  \multicolumn{1}{r|}{244.53 $\pm$ 7.94} &
  190.83 $\pm$ 26.97 \\ \hline
\multicolumn{1}{|c|}{1} &
  70 &
  \multicolumn{1}{r|}{-18.47 $\pm$ 1.30} &
  -18.88 $\pm$ 1.08 &
  \multicolumn{1}{r|}{243.08 $\pm$ 23.85} &
  186.83 $\pm$ 19.25 \\ \hline
\multicolumn{1}{|c|}{1} &
  60 &
  \multicolumn{1}{r|}{-19.89 $\pm$ 0.05} &
  -20.16 $\pm$ 0.09 &
  \multicolumn{1}{r|}{232.95 $\pm$ 17.10} &
  214.58 $\pm$ 13.55 \\ \hline
\multicolumn{1}{|c|}{1} &
  50 &
  \multicolumn{1}{r|}{-19.63 $\pm$ 0.54} &
  -19.80 $\pm$ 0.56 &
  \multicolumn{1}{r|}{251.46 $\pm$ 33.61} &
  190.62 $\pm$ 22.40 \\ \hline
\multicolumn{1}{|c|}{1} &
  40 &
  \multicolumn{1}{r|}{-19.93 $\pm$ 0.04} &
  -20.19 $\pm$ 0.11 &
  \multicolumn{1}{r|}{218.78 $\pm$ 2.93} &
  163.42 $\pm$ 14.12 \\ \hline
\multicolumn{1}{|c|}{1} &
  30 &
  \multicolumn{1}{r|}{-19.95 $\pm$ 0.03} &
  -20.17 $\pm$ 0.17 &
  \multicolumn{1}{r|}{221.73 $\pm$ 4.20} &
  170.33 $\pm$ 21.35 \\ \hline
\multicolumn{1}{|c|}{1} &
  20 &
  \multicolumn{1}{r|}{-19.75 $\pm$ 0.32} &
  -20.11 $\pm$ 0.25 &
  \multicolumn{1}{r|}{244.51 $\pm$ 4.90} &
  203.97 $\pm$ 23.26 \\ \hline
\multicolumn{1}{|c|}{1} &
  10 &
  \multicolumn{1}{r|}{-19.87 $\pm$ 0.07} &
  -20.23 $\pm$ 0.06 &
  \multicolumn{1}{r|}{238.90$\pm$ 15.14} &
  188.23 $\pm$ 32.58 \\ \hline
\rowcolor[HTML]{EFEFEF} 
\multicolumn{1}{|c|}{\cellcolor[HTML]{EFEFEF}2} &
  80 &
  \multicolumn{1}{r|}{\cellcolor[HTML]{9AFF99}7.15 $\pm$ 20.80} &
  \cellcolor[HTML]{9AFF99}6.64 $\pm$ 21.14 &
  \multicolumn{1}{r|}{\cellcolor[HTML]{9AFF99}444.18 $\pm$ 40.78} &
  341.47 $\pm$ 25.71 \\ \hline
\rowcolor[HTML]{EFEFEF} 
\multicolumn{1}{|c|}{\cellcolor[HTML]{EFEFEF}2} &
  70 &
  \multicolumn{1}{r|}{\cellcolor[HTML]{EFEFEF}-6.68 $\pm$ 22.92} &
  -6.99 $\pm$ 22.80 &
  \multicolumn{1}{r|}{\cellcolor[HTML]{EFEFEF}371.68 $\pm$ 58.30} &
  296.10 $\pm$ 36.34 \\ \hline
\rowcolor[HTML]{EFEFEF} 
\multicolumn{1}{|c|}{\cellcolor[HTML]{EFEFEF}2} &
  60 &
  \multicolumn{1}{r|}{\cellcolor[HTML]{EFEFEF}-19.86 $\pm$ 0.05} &
  -20.27 $\pm$ 0.18 &
  \multicolumn{1}{r|}{\cellcolor[HTML]{EFEFEF}371.21 $\pm$ 6.30} &
  \cellcolor[HTML]{9AFF99}350.15 $\pm$ 10.13 \\ \hline
\rowcolor[HTML]{EFEFEF} 
\multicolumn{1}{|c|}{\cellcolor[HTML]{EFEFEF}2} &
  50 &
  \multicolumn{1}{r|}{\cellcolor[HTML]{EFEFEF}-16.72 $\pm$ 5.48} &
  -17.17 $\pm$ 5.20 &
  \multicolumn{1}{r|}{\cellcolor[HTML]{EFEFEF}283.81 $\pm$ 76.94} &
  217.58$\pm$ 62.06 \\ \hline
\rowcolor[HTML]{EFEFEF} 
\multicolumn{1}{|c|}{\cellcolor[HTML]{EFEFEF}2} &
  40 &
  \multicolumn{1}{r|}{\cellcolor[HTML]{EFEFEF}-19.91 $\pm$ 0.03} &
  -20.19 $\pm$ 0.07 &
  \multicolumn{1}{r|}{\cellcolor[HTML]{EFEFEF}252.58 $\pm$ 22.31} &
  194.52 $\pm$ 26.93 \\ \hline
\rowcolor[HTML]{EFEFEF} 
\multicolumn{1}{|c|}{\cellcolor[HTML]{EFEFEF}2} &
  30 &
  \multicolumn{1}{r|}{\cellcolor[HTML]{EFEFEF}-17.24 $\pm$ 2.34} &
  -17.52 $\pm$ 2.43 &
  \multicolumn{1}{r|}{\cellcolor[HTML]{EFEFEF}248.98 $\pm$ 5.78} &
  208.03 $\pm$ 7.23 \\ \hline
\rowcolor[HTML]{EFEFEF} 
\multicolumn{1}{|c|}{\cellcolor[HTML]{EFEFEF}2} &
  20 &
  \multicolumn{1}{r|}{\cellcolor[HTML]{EFEFEF}-19.87 $\pm$ 0.07} &
  -20.17 $\pm$ 0.16 &
  \multicolumn{1}{r|}{\cellcolor[HTML]{EFEFEF}241.50 $\pm$ 10.81} &
  190.52 $\pm$ 31.40 \\ \hline
\rowcolor[HTML]{EFEFEF} 
\multicolumn{1}{|c|}{\cellcolor[HTML]{EFEFEF}2} &
  10 &
  \multicolumn{1}{r|}{\cellcolor[HTML]{EFEFEF}-19.91 $\pm$ 0.06} &
  -20.24 $\pm$ 0.09 &
  \multicolumn{1}{r|}{\cellcolor[HTML]{EFEFEF}243.51 $\pm$ 5.16} &
  209.43 $\pm$ 32.67 \\ \hline
\multicolumn{1}{|c|}{3} &
  40 &
  \multicolumn{1}{r|}{\cellcolor[HTML]{9AFF99}20.06 $\pm$ 0.44} &
  \cellcolor[HTML]{9AFF99}19.82 $\pm$ 0.88 &
  \multicolumn{1}{r|}{\cellcolor[HTML]{9AFF99}596.50 $\pm$ 182.35} &
  \cellcolor[HTML]{9AFF99}544.43 $\pm$ 166.79 \\ \hline
\multicolumn{1}{|c|}{3} &
  30 &
  \multicolumn{1}{r|}{\cellcolor[HTML]{FFFFFF}-10.99 $\pm$ 10.42} &
  \cellcolor[HTML]{FFFFFF}-11.80 $\pm$ 9.05 &
  \multicolumn{1}{r|}{\cellcolor[HTML]{FFFFFF}274.38 $\pm$ 5.24} &
  \cellcolor[HTML]{FFFFFF}212.07 $\pm$ 34.62 \\ \hline
\multicolumn{1}{|c|}{3} &
  20 &
  \multicolumn{1}{r|}{\cellcolor[HTML]{FFFFFF}-19.92 $\pm$ 0.01} &
  \cellcolor[HTML]{FFFFFF}-20.12 $\pm$ 0.10 &
  \multicolumn{1}{r|}{\cellcolor[HTML]{FFFFFF}293.12 $\pm$ 28.46} &
  \cellcolor[HTML]{FFFFFF}228.63 $\pm$ 28.06 \\ \hline
\multicolumn{1}{|c|}{3} &
  10 &
  \multicolumn{1}{r|}{\cellcolor[HTML]{FFFFFF}-19.93 $\pm$ 0.03} &
  \cellcolor[HTML]{FFFFFF}-20.35 $\pm$ 0.39 &
  \multicolumn{1}{r|}{\cellcolor[HTML]{FFFFFF}251.60 $\pm$ 12.27} &
  \cellcolor[HTML]{FFFFFF}194.13 $\pm$ 31.28 \\ \hline
\rowcolor[HTML]{9AFF99} 
\multicolumn{1}{|c|}{\cellcolor[HTML]{EFEFEF}4} &
  \cellcolor[HTML]{EFEFEF}20 &
  \multicolumn{1}{r|}{\cellcolor[HTML]{9AFF99}{\color[HTML]{000000} -14.86 $\pm$ 5.39}} &
  {\color[HTML]{000000} -15.77 $\pm$ 4.82} &
  \multicolumn{1}{r|}{\cellcolor[HTML]{9AFF99}{\color[HTML]{000000} 439.72 $\pm$ 26.27}} &
  {\color[HTML]{000000} 378.00 $\pm$ 28.70} \\ \hline
\rowcolor[HTML]{EFEFEF} 
\multicolumn{1}{|c|}{\cellcolor[HTML]{EFEFEF}4} &
  10 &
  \multicolumn{1}{r|}{\cellcolor[HTML]{EFEFEF}-19.86 $\pm$ 0.05} &
  -20.23 $\pm$ 0.08 &
  \multicolumn{1}{r|}{\cellcolor[HTML]{EFEFEF}297.45 $\pm$ 9.98} &
  252.77 $\pm$ 13.72 \\ \hline
\end{tabular}
\caption{Training and testing performance of every glimpse size for every number of patches tested in Pong and SpaceInvaders.}
\label{tab:glimpse_study_atari}
\end{table}

\begin{table}[ht]
\centering
\begin{tabular}{cc|rr|}
\cline{3-4}
\multicolumn{1}{l}{} &
  \multicolumn{1}{l|}{} &
  \multicolumn{2}{c|}{CarRacing-v0} \\ \hline
\rowcolor[HTML]{AFAFAF} 
\multicolumn{1}{|c|}{\cellcolor[HTML]{AFAFAF}No. Patches} &
  Glimpse Size &
  \multicolumn{1}{c|}{\cellcolor[HTML]{AFAFAF}Max. Train Avg.} &
  \multicolumn{1}{c|}{\cellcolor[HTML]{AFAFAF}Test Avg.} \\ \hline
\multicolumn{1}{|c|}{1} &
  96 &
  \multicolumn{1}{r|}{\cellcolor[HTML]{9AFF99}815.12 $\pm$ 5.75} &
  \cellcolor[HTML]{9AFF99}660.02 $\pm$ 69.38 \\ \hline
\multicolumn{1}{|c|}{1} &
  90 &
  \multicolumn{1}{r|}{675.12 $\pm$ 84.89} &
  607.46 $\pm$ 39.81 \\ \hline
\multicolumn{1}{|c|}{1} &
  80 &
  \multicolumn{1}{r|}{741.03 $\pm$ 112.73} &
  534.56 $\pm$ 72.77 \\ \hline
\multicolumn{1}{|c|}{1} &
  70 &
  \multicolumn{1}{r|}{747.12 $\pm$ 103.91} &
  258.70 $\pm$ 274.03 \\ \hline
\multicolumn{1}{|c|}{1} &
  60 &
  \multicolumn{1}{r|}{692.40 $\pm$ 121.21} &
  546.35 $\pm$ 40.01 \\ \hline
\multicolumn{1}{|c|}{1} &
  50 &
  \multicolumn{1}{r|}{559.05 $\pm$ 19.45} &
  361.30 $\pm$ 73.27 \\ \hline
\multicolumn{1}{|c|}{1} &
  40 &
  \multicolumn{1}{r|}{539.68 $\pm$ 11.35} &
  485.85 $\pm$ 9.80 \\ \hline
\multicolumn{1}{|c|}{1} &
  30 &
  \multicolumn{1}{r|}{187.67 $\pm$ 136.04} &
  32.64 $\pm$ 81.82 \\ \hline
\multicolumn{1}{|c|}{1} &
  20 &
  \multicolumn{1}{r|}{155.15 $\pm$ 77.05} &
  152.99 $\pm$ 75.43 \\ \hline
\multicolumn{1}{|c|}{1} &
  10 &
  \multicolumn{1}{r|}{138.51 $\pm$ 88.40} &
  128.90 $\pm$ 94.17 \\ \hline
\rowcolor[HTML]{EFEFEF} 
\multicolumn{1}{|c|}{\cellcolor[HTML]{EFEFEF}2} &
  48 &
  \multicolumn{1}{r|}{\cellcolor[HTML]{9AFF99}748.70 $\pm$ 74.90} &
  544.58 $\pm$ 132.04 \\ \hline
\rowcolor[HTML]{EFEFEF} 
\multicolumn{1}{|c|}{\cellcolor[HTML]{EFEFEF}2} &
  40 &
  \multicolumn{1}{r|}{\cellcolor[HTML]{EFEFEF}694.50 $\pm$ 107.94} &
  \cellcolor[HTML]{9AFF99}641.11 $\pm$ 57.42 \\ \hline
\rowcolor[HTML]{EFEFEF} 
\multicolumn{1}{|c|}{\cellcolor[HTML]{EFEFEF}2} &
  30 &
  \multicolumn{1}{r|}{\cellcolor[HTML]{EFEFEF}616.74 $\pm$ 84.43} &
  347.20 $\pm$ 243.31 \\ \hline
\rowcolor[HTML]{EFEFEF} 
\multicolumn{1}{|c|}{\cellcolor[HTML]{EFEFEF}2} &
  20 &
  \multicolumn{1}{r|}{\cellcolor[HTML]{EFEFEF}465.76 $\pm$ 74.51} &
  375.30 $\pm$ 164.83 \\ \hline
\rowcolor[HTML]{EFEFEF} 
\multicolumn{1}{|c|}{\cellcolor[HTML]{EFEFEF}2} &
  10 &
  \multicolumn{1}{r|}{\cellcolor[HTML]{EFEFEF}161.44 $\pm$ 36.96} &
  145.80 $\pm$ 48.40 \\ \hline
\multicolumn{1}{|c|}{3} &
  24 &
  \multicolumn{1}{r|}{\cellcolor[HTML]{9AFF99}700.53 $\pm$ 28.06} &
  \cellcolor[HTML]{FFFFFF}472.35 $\pm$ 295.73 \\ \hline
\multicolumn{1}{|c|}{3} &
  20 &
  \multicolumn{1}{r|}{\cellcolor[HTML]{FFFFFF}676.82 $\pm$ 84.89} &
  \cellcolor[HTML]{9AFF99}564.00 $\pm$ 56.42 \\ \hline
\multicolumn{1}{|c|}{3} &
  10 &
  \multicolumn{1}{r|}{\cellcolor[HTML]{FFFFFF}545.18 $\pm$ 92.57} &
  \cellcolor[HTML]{FFFFFF}509.85 $\pm$ 70.74 \\ \hline
\end{tabular}
\caption{Training and testing performance of every glimpse size for every number of patches tested in CarRacing.}
\label{tab:glimpse_study_carracing}
\end{table}

\newpage

\section{Algorithms hyperparameters}

\begin{table}[ht]
\centering
\begin{tabular}{c|l|c|}
\cline{2-3}
\multicolumn{1}{l|}{}                          & \multicolumn{1}{c|}{\textbf{Hyperparameter}} & \textbf{Value(s)}  \\ \hline
\multicolumn{1}{|c|}{\multirow{21}{*}{\rotatebox[origin=c]{90}{PPO}}}    & Advantage normalization                      & True               \\ \cline{2-3} 
\multicolumn{1}{|c|}{} & Annealing learning rate          & True             \\ \cline{2-3} 
\multicolumn{1}{|c|}{} & Batch size                       & {[}1024, 2048{]} \\ \cline{2-3} 
\multicolumn{1}{|c|}{}                         & Clipping coefficient - action                & {[}0.1, 0.2{]}     \\ \cline{2-3} 
\multicolumn{1}{|c|}{} & Clipping coefficient - location  & 0.2              \\ \cline{2-3} 
\multicolumn{1}{|c|}{} & Clipped value loss               & True             \\ \cline{2-3} 
\multicolumn{1}{|c|}{} & Entropy coefficient              & {[}0.01, 0{]}    \\ \cline{2-3} 
\multicolumn{1}{|c|}{} & GAE lambda                       & 0.95             \\ \cline{2-3} 
\multicolumn{1}{|c|}{} & Gamma                            & 0.99             \\ \cline{2-3} 
\multicolumn{1}{|c|}{} & Grayscale                        & True             \\ \cline{2-3} 
\multicolumn{1}{|c|}{}                         & Learning rate - action                       & {[}2.5e-4, 3e-4{]} \\ \cline{2-3} 
\multicolumn{1}{|c|}{} & Learning rate - location         & 3e-5             \\ \cline{2-3} 
\multicolumn{1}{|c|}{} & Locator normal distrib. variance & 0.1              \\ \cline{2-3} 
\multicolumn{1}{|c|}{} & Maximum gradient clipping norm   & 0.5              \\ \cline{2-3} 
\multicolumn{1}{|c|}{} & Minibatch size                   & {[}256, 64{]}    \\ \cline{2-3} 
\multicolumn{1}{|c|}{} & No. environments                 & {[}8, 1{]}       \\ \cline{2-3} 
\multicolumn{1}{|c|}{} & No. minibatches                  & {[}4, 32{]}      \\ \cline{2-3} 
\multicolumn{1}{|c|}{} & No. steps                        & {[}128, 2048{]}  \\ \cline{2-3} 
\multicolumn{1}{|c|}{} & Optimizer                        & Adam             \\ \cline{2-3}
\multicolumn{1}{|c|}{} & Update k epochs                  & {[}4, 10{]}      \\ \cline{2-3} 
\multicolumn{1}{|c|}{} & Value function coefficient       & 0.5              \\ \hline
\multicolumn{1}{|l|}{\multirow{5}{*}{\rotatebox[origin=c]{90}{Glimpse}}} & Glimpse scale                                & 2                  \\ \cline{2-3} 
\multicolumn{1}{|l|}{} & Glimpse FC layer size            & {[}384, 512{]}   \\ \cline{2-3} 
\multicolumn{1}{|l|}{} & Location FC layer size           & 256              \\ \cline{2-3} 
\multicolumn{1}{|l|}{} & LSTM size                        & 128              \\ \cline{2-3} 
\multicolumn{1}{|l|}{} & No. glimpses                     & 1                \\ \hline
\end{tabular}
\caption{List of parameters used in GBAC, in the Atari games and CarRacing, respectively}
\label{tab:hyperparameters_gbac}
\end{table}

\end{document}